\documentclass{article}

    \usepackage[final]{neurips_2024}

\usepackage[utf8]{inputenc} % allow utf-8 input
\usepackage[T1]{fontenc}    % use 8-bit T1 fonts
\usepackage{hyperref}       % hyperlinks
\usepackage{url}            % simple URL typesetting
\usepackage{booktabs}       % professional-quality tables
\usepackage{amsfonts}       % blackboard math symbols
\usepackage{nicefrac}       % compact symbols for 1/2, etc.
\usepackage{microtype}      % microtypography

\usepackage[table]{xcolor}
\usepackage{amssymb}
\usepackage{amsmath}
\usepackage[pdftex]{graphicx}

\usepackage{algorithmic}
\usepackage{algorithm}

\DeclareMathOperator*{\argmin}{arg\,min}

\title{Dynamic Subset Tuning: Expanding the Operational Range of Parameter-Efficient Training for Large Language Models}

\author{%
  Felix Stahlberg, Jared Lichtarge, Shankar Kumar\\
  Google Research \\
  \texttt{\{fstahlberg,lichtarge,shankarkumar\}@google.com} \\
}

\begin{document}

\maketitle

\begin{abstract}
  We propose a novel parameter-efficient training (PET) method for large language models that adapts models to downstream tasks by optimizing a small subset of the existing model parameters. Unlike prior methods, this subset is not fixed in location but rather which parameters are modified evolves over the course of training. This dynamic parameter selection can yield good performance with many fewer parameters than extant methods. Our method enables a seamless scaling of the subset size across an arbitrary proportion of the total model size, while popular PET approaches like prompt tuning and LoRA cover only a small part of this spectrum. We match or outperform prompt tuning and LoRA in most cases on a variety of NLP tasks (MT, QA, GSM8K, SuperGLUE) for a given parameter budget across different model families and sizes. 
\end{abstract}

\section{Introduction}

Large language models (LLMs) such as GPT-4 \citep{gpt4} and PaLM \citep{palm2,palm} have demonstrated remarkable performance on various natural language processing (NLP) tasks. A common paradigm for adapting LLMs to specific NLP tasks is fine-tuning the pre-trained model on a downstream task dataset. Effective regularization is paramount since fine-tuning is often prone to overfitting due to the large model size and the small number of examples in the fine-tuning dataset. Furthermore, storing a full model version for each task becomes prohibitive as the number of tasks grows. Parameter-efficient training (PET)~\citep{delta-tuning-overview} is a family of approaches that regularizes fine-tuning by allowing only a small number of parameters to change, thus reducing the storage overhead for each task drastically. Popular techniques such as prompt tuning \citep{pt} and LoRA \citep{lora} fall into this category.

A recent line of work in PET keeps the internal structure of the seed model intact by identifying a small subset of parameters that is optimized in fine-tuning while freezing all others. \citet{delta-tuning-overview} refer to this as {\em specification-based delta tuning}. Typically, these approaches specify the free parameter set {\em once} either before or after model adaptation. The subset is chosen based on either heuristics \citep{ben-zaken-etal-2022-bitfit}, Fisher information \citep{child-tuning,fixed-mask}, or the fine-tuning training signal \citep{piggyback,diff-pruning,ansell-etal-2022-composable,skilllocalization}.

In this work, we propose a novel algorithm called {\em dynamic subset tuning} (DST) that efficiently re-selects the set of free parameters {\em in each training step} without a substantial slowdown in training. Our method is more flexible than prompt tuning and LoRA as it allows us to precisely specify the fine-tuning parameter budget as a fraction of the model size. Our method often matches or outperforms prompt tuning and LoRA with comparable parameter budget on various NLP tasks, and covers a much wider range of subset sizes. In particular, it enables the use of PET with many fewer free parameters (down to 0.00001\% of the network) than most prior work, allowing DST to make use of even very small training sets.

\section{Dynamic subset tuning}
\label{sec:sdt}

Let $\Theta^{(t)}\in\mathbb{R}^n$ be the parameter vector at time step $t$, where $n$ is the number of parameters in the model. $\Theta^{(0)}\in\mathbb{R}^n$ refers to the parameters of the pre-trained seed model. In DST, the fine-tuned parameter vector differs from the seed model in exactly $\epsilon n$ components ($\epsilon\in (0,1)$). This constraint can be enforced at different granularity levels, 
which we describe using a partition $\mathcal{S}$ of $[1,n]$ into $k$ ``silos''. $\mathcal{S} = \{S_i\}_{i=1}^k$ where each $S_i \subseteq [1,n]$, and $\bigcup_{i=1}^k S_i = [1,n]$, for any $i \ne j$, $S_i \bigcap S_j =\emptyset$, and $\sum_{i=1}^k |S_i| = n$
(i.e.\ a mutually exclusive and jointly exhaustive division of the model parameters). In this paper, we use \textbf{per-module-and-layer siloing}: Each module in each layer is a separate silo that has a constant fraction of free parameters, but the set of free parameters can still be spread out within the silo (e.g.\ within a weight matrix). Appendix \ref{sec:dist-and-silo} compares this siloing strategy with alternative approaches.
We denote the set of valid parameter vectors as $\mathcal{H}_{\epsilon,\mathcal{S}}$:
% \begin{minipage}{1.0\linewidth}
\begin{equation}
    \mathcal{H}_{\epsilon,\mathcal{S}} :=  \{ \Theta\in \mathbb{R}^n| \forall S\in\mathcal{S}: \epsilon |S| = \sum_{i\in S} I(\Theta^{(0)}_i\neq\Theta_i)\}.
\end{equation}
% \end{minipage}
where $\Theta_i$ is the $i^{\text{th}}$ component of the parameter vector $\Theta$ and $I(\cdot)$ is the indicator function (1 if the condition is true, 0 otherwise). The $\epsilon$-constraint is enforced in each silo separately:
\begin{minipage}{1.0\linewidth}
\begin{equation}
\label{eq:eps}
    \forall t\in \mathbb{N}: \Theta^{(t)}\in\mathcal{H}_{\epsilon,\mathcal{S}}.
\end{equation}
\end{minipage}
The silo-level $\epsilon$-constraints in Eq.\ \ref{eq:eps} imply that, at each time step $t$, the number of parameters across the {\em full} model that differ from $\Theta^{(0)}$ is exactly $\epsilon n$ since $\mathcal{H}_{\epsilon,\mathcal{S}}\subset \mathcal{H}_{\epsilon,\{[1,n]\}}$. We denote the updates computed using the optimizer's update rule with $u^{(t)}\in\mathbb{R}^n$ such that a full update step to $\hat{\Theta}^{(t+1)}\in\mathbb{R}^n$ can be described as:

\begin{minipage}{1.0\linewidth}
\begin{equation}
    \hat{\Theta}^{(t+1)}=\Theta^{(t)}+u^{(t)}.
\end{equation}

\end{minipage}
Vanilla fine-tuning would use $\hat{\Theta}^{(t+1)}$ directly in the next training iteration ($\Theta^{(t+1)}\overset{\text{\tiny{Full-FT}}}{=}\hat{\Theta}^{(t+1)}$). In DST, however, we aim to find a $\Theta^{(t+1)}$ that is {\em close} to the fully updated $\hat{\Theta}^{(t+1)}$ but satisfies Eq.~\ref{eq:eps}:
\begin{equation}
\label{eq:argmin}
    \Theta^{(t+1)}=\argmin_{\Theta\in\mathcal{H}_{\epsilon,\mathcal{S}}} \sum_{i=1}^n d(\Theta_i, \hat{\Theta}^{(t+1)}_i, \Theta^{(0)}_i),
\end{equation}
where $d(\cdot,\cdot, \cdot)$ is a distance function. An example is the component-wise \textbf{inverse-relative distance function} that biases towards both large differences and seed parameter magnitudes:

\begin{minipage}{1.0\linewidth}
\begin{equation}
   d(\Theta_i, \hat{\Theta}^{(t+1)}_i, \Theta^{(0)}_i) =|\Theta^{(0)}_i(\Theta_i-\hat{\Theta}^{(t+1)}_i)|
\end{equation}
\end{minipage}

Appendix \ref{sec:dist-and-silo} compares different variants of this distance function.
Eq.\ \ref{eq:argmin} can be solved for each silo $S\in\mathcal{S}$ separately by using values from $\hat{\Theta}^{(t+1)}$ in the top-$\epsilon|S|$ positions of the distance vector between $\Theta^{(0)}$ and $\hat{\Theta}^{(t+1)}$, and resetting the remaining values to $\Theta^{(0)}$. More formally, let $\Delta$ be the distance vector and $q_S\in\mathbb{R}_+$ be the threshold that separates the top-$\epsilon|S|$ distances from the rest:

\begin{minipage}{1.0\linewidth}

\begin{equation}
    \forall i\in [1,n]: \Delta_i = d(\Theta^{(0)}_i, \hat{\Theta}^{(t+1)}_i, \Theta^{(0)}_i)\text{ and } \epsilon|S| = \sum_{i\in S} I(\Delta_i> q_S).
\end{equation}
We construct the final parameter vector $\Theta^{(t+1)}\big|_S$ for silo $S$ as follows:
\begin{equation}
    \forall i\in S: \Theta^{(t+1)}_i = \begin{cases}
			\hat{\Theta}^{(t+1)}_i, & \text{if }\Delta_i > q_S\\
            \Theta^{(0)}_i, & \text{otherwise}
		 \end{cases}.
\end{equation}
\end{minipage}

Alg.\ \ref{alg:sdt} shows a full update step in DST. \footnote{In practice, we use an iterative approximation to the \texttt{quantile}() function in line 5 which makes the computational overhead negligible. See Appendix \ref{sec:quantile-computation} for more details.}

\begin{algorithm}[t!]
\small
\caption{\small DST update function for computing the model parameters $\Theta^{(t+1)}$ for the next training iteration. Note that \texttt{compute\_full\_updates()} may have additional dependencies such as the optimizer state in momentum-based optimizers that we left out for the sake of simplicity.}
\label{alg:sdt}
\begin{algorithmic}[1]
\REQUIRE{$\Theta^{(0)}$: Seed params; $\Theta^{(t)}$: Params at time step $t$; $\epsilon$: Fraction of free params; $\mathcal{S}$: Siloing partition}
% \REQUIRE{$\Theta^{(0)}$: Seed parameters at time step $0$.}
% \REQUIRE{$\Theta^{(t)}$: Parameters at time step $t$.}
% \REQUIRE{$\epsilon$: Fraction of free parameters.}
% \REQUIRE{$\mathcal{S}$: Siloing partition.}
\STATE{$u\gets \mathtt{compute\_full\_updates}(\Theta^{(t)})$} \COMMENT{Apply optimizer's update rule}
\STATE{$\hat{\Theta} \gets \Theta^{(t)} + u$}
\STATE{$\Delta \gets d(\Theta^{(0)}, \hat{\Theta}, \Theta^{(0)}) $} \COMMENT{Component-wise}
\FOR{$S\in\mathcal{S}$}
\STATE{$q\gets\mathtt{quantile}(\Delta\big|_S, 1-\epsilon)$}
\STATE{$\Theta^{(t+1)}\big|_S\gets\mathtt{where}(\Delta\big|_S>q, \hat{\Theta}\big|_S, \Theta^{(0)}\big|_S)$}
\ENDFOR
\RETURN{$\Theta^{(t+1)}$}
\end{algorithmic}
\end{algorithm}

\section{Experimental setup}

All our training runs start off from publicly available pre-trained checkpoints. For the foundation model PaLM 2 \citep{palm2} we use the three sizes available via the Google Cloud API: \textbf{Gecko}, \textbf{Otter}, and \textbf{Bison}. For T5 we use the T5 1.1 checkpoints\footnote{\url{https://github.com/google-research/text-to-text-transfer-transformer/blob/main/released_checkpoints.md}} available in T5X \citep{t5x}: \textbf{Small} (77M parameters), \textbf{Base} (250M parameters), \textbf{Large} (800M parameters), \textbf{XL} (3B parameters), and \textbf{XXL} (11B parameters). AdaFactor \citep{adafactor} state variables like momentum are computed using the fully updated parameters $\hat{\Theta}^{(t)}$, not the post-processed ones $\Theta^{(t)}$. The dropout rates and learning rates are tuned on the respective validation sets -- see Appendix \ref{sec:training-hp} for details. We test our method on the following benchmarks: Machine Translation (WMT22 German-English), Trivia-QA (closed-book, open-domain question answering), GSM8K (math problems), SuperGLUE (eight language understanding tasks). Appendix \ref{sec:data} lists more details about the datasets and our evaluation protocols.

\section{Results}
\label{sec:results}

\begin{figure}[t!]
\centering
\small
\begin{tabular}{@{\hspace{0em}}c@{\hspace{0em}}c@{\hspace{0em}}}
\includegraphics[scale=0.28]{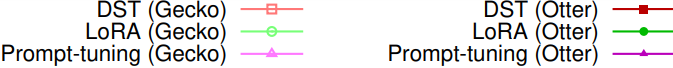} & \includegraphics[scale=0.9]{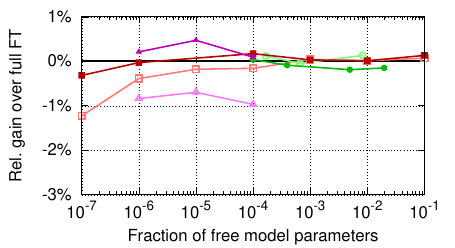} \\
& (a) Machine translation (German-English. WMT) \vspace{0.3em} \\
\includegraphics[scale=0.9]{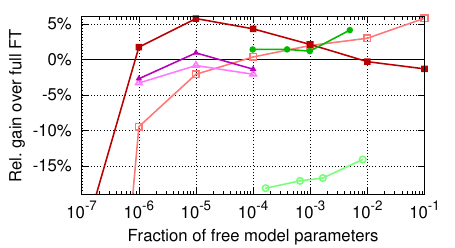} & \includegraphics[scale=0.9]{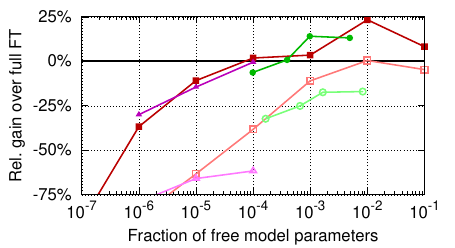} \\
(b) TriviaQA \vspace{0.3em} & (c) GSM8K \vspace{0.3em} \\
\includegraphics[scale=0.9]{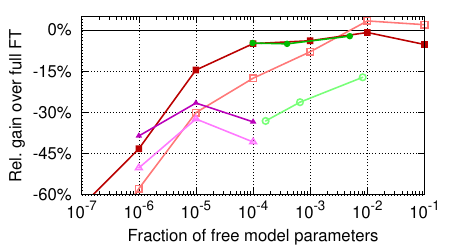} & \includegraphics[scale=0.9]{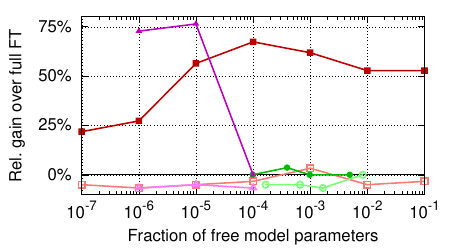}\\
(d) SuperGLUE \vspace{0.3em} & (e) COPA \\
\end{tabular}

\caption{Validation set performance of the Gecko and Otter models as a function of the fraction of free parameters relative to full fine-tuning.}
\label{fig:main-eval-dev-rel}
\end{figure}

\subsection{Comparison with prompt tuning and LoRA}

We compare DST with two popular PET baselines: \textbf{Prompt tuning} \citep{pt} is an extension of prompt engineering. It replaces hard-coded textual prompts with soft prompt-embeddings that are trained on the downstream task. \textbf{LoRA} \citep{lora} adds linear bottleneck structures to the network that are parallel to the attention weight matrices. All parameters besides the ones in the added bottlenecks are frozen during fine-tuning.

Fig.\ \ref{fig:main-eval-dev-rel} compares both PET methods with DST as a function of the number of free parameters. We show the performance of the PET methods relative to the respective fully fine-tuned model.\footnote{Appendix \ref{sec:main-eval-abs} lists the absolute scores of all systems.} We vary the prompt length in prompt tuning between 1 and 100. LoRA ranks range between 1 and 50. Like \citet{lora} we noticed that higher LoRA ranks often do not yield further gains, possibly due to regularization being less effective. Similarly, prompt tuning peaks at a prompt length of 10 ($\approx 10^{-5}$ free parameter fraction). Compared to LoRA and prompt-tuning, DST curves (red) show good performance across a broader range of the allocated parameter budget. This shows that there is more flexibility in controlling the parameter budget. On very small training sets (Fig.\ \ref{fig:main-eval-dev-rel}e), DST with the Otter model results in a smooth performance curve, while other systems are either erratic (prompt tuning) or stay close to the chance level.

\subsection{T5 experiments}

\begin{figure}[t!]
\centering
\small
\begin{minipage}{0.48\linewidth}
\includegraphics[scale=0.86]{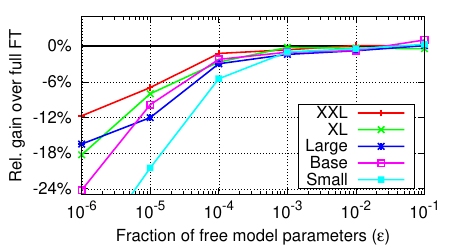}
\caption{Average scores of T5 models on SuperGLUE compared to full fine-tuning as a function of $\epsilon$.}
\label{fig:t5}
% \end{figure}
\end{minipage}\hspace{0.3cm}
\begin{minipage}{0.48\linewidth}
% \begin{figure}[t!]
% \centering
% \small
\includegraphics[scale=0.86]{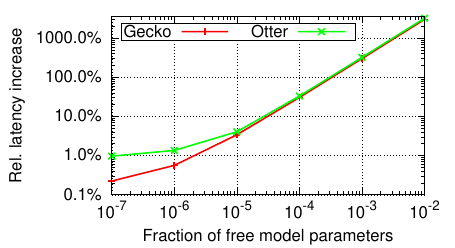}
\caption{Latency increase of applying subset weights on-the-fly before decoding as a function of $\epsilon$. We use the MT test sentence ``Wie geht es dir heute?''.}
\label{fig:latency}
\end{minipage}
\end{figure}

Fig.\ \ref{fig:t5} shows the trade-off between the quality and the subset size for different T5 model sizes. For $\epsilon\geq 10^{-4}$ (i.e. 0.01\% free parameters) the performance of all our T5 models except {\em Small} stays within 3\% of the full fine-tuning baseline. Smaller models are more affected by an even lower $\epsilon$ than larger models. This suggests that the system performance mainly depends on the absolute size of the subset: {\em Small} is more than 100x smaller than {\em XXL}, and so are the absolute sizes of their subsets for the same~$\epsilon$.
%We report minor gains over full fine-tuning for $\epsilon=10^{-1}$.

\subsection{Multi-subset serving latency}

In addition to regularization, a common use case for PET is model sharing. For example, in DST, a single base model can be adapted to the task at hand by swapping out the subset at inference time. Fig.\ \ref{fig:latency} shows that the latency increase of applying DST parameter subsets on-the-fly is close or below 1\% for $\epsilon \leq 10^{-6}$ for a basic JAX-based implementation on a 2x2 TPU v5e, potentially making network-based client-server scenarios viable in a scenario where the subset is sent by the client. Our multi-subset inference implementation is not suitable for efficiently serving subsets larger than $\epsilon = 10^{-4}$, but better support for sparse matrix operations\footnote{\url{https://jax.readthedocs.io/en/latest/jax.experimental.sparse.html}} may improve the efficiency in the future. Note that there is no latency increase for single-subset serving as the subset can be embedded in the model at startup time.

\section{Related work}

Besides LoRA and prompt tuning, DST is closely related to delta tuning \citep{delta-tuning-overview} approaches such as \textit{diff pruning} \citep{piggyback,diff-pruning}. Diff pruning uses a continuous relaxation of the subset constraint which is determinized {\em once} after training. Other delta tuning methods choose the subnetwork mask {\em prior} to fine-tuning based on the Fisher information \citep{child-tuning,fixed-mask}. In contrast, DST re-selects the subset in every step. DST is more flexible than heuristic-based delta tuning approaches like BitFit \citep{ben-zaken-etal-2022-bitfit} since the subset is learnt. Our absolute distance function in Table \ref{tab:d} is inspired by \citet{ansell-etal-2022-composable}. We extend prior work on delta tuning by (a) proposing an algorithm that allows for discrete subset churn at each training step, (b) introducing the concept of siloing, and (c) comparing distance normalizing strategies. We show on a number of benchmarks that DST covers a wider range of subset sizes than our baselines.

DST uses the insight that a small subnetwork is sufficient for adapting a model to a downstream task. This is connected to the lottery ticket hypothesis \citep{lth} that states that subnetworks can reach test accuracy comparable to the original network. Similarly, \citet{zhang-etal-2022-moefication} suggested that the activations in vanilla Transformers tend to be sparse.
%and resemble internal mixture-of-expert structures.

Delta tuning approaches such as DST are related to model pruning. Similar to movement pruning \citep{movement-pruning}, our distance functions in Table \ref{tab:d} rely on changes in weights. Our inverse-relative distance function is motivated by magnitude pruning \citep{magnitude-pruning}. Although intuitively similar, DST's goal is PET and not model shrinking.

Adapters \citep{NIPS2017_e7b24b11,adapter,pfeiffer-etal-2021-adapterfusion,he-etal-2021-effectiveness} are small bottleneck layers inserted into the pre-trained model that are learnt in fine-tuning while the rest of the model is frozen. Unlike adapters, DST leaves the network structure untouched.

\citet{su-etal-2023-exploring} and Xu and Zhang \citep{randommasking} use random masks to reduce the number of free parameters in fine-tuning. We compare DST with random masking in Sec. \ref{sec:dist-and-silo}. \citet{compressinglora} reduce the number of parameters in LoRA by factoring the low rank matrices using random basis matrices. By contrast, DST does not require a low rank factorization.

We demonstrated that DST has a regularization effect that yields gains over full fine-tuning when training data is scarse. This confirms prior work \citep{vu-etal-2022-spot,NEURIPS2022_0cde695b,pmlr-v202-oymak23a} showing that PET can outperform full fine-tuning. Regularizing training by limiting updates to a subset of the model parameters has also been explored by \citet{mixout} and \citet{dps}. Similar ideas inspired techniques like block coordinate descent or stacking \citep{stacking} that aim to improve the efficiency and scalability of training large models. In these methods, unlike DST, parameters are not reset to their seed value when they drop out of the subset, resulting in a final model that differs more from the seed and has a larger effective subset size. Elastic weight consolidation \citep[EWC]{ewc} and DST both aim to regularize by keeping parameters close to their seed values, but EWC allows {\em all} parameters to change. DST is less expensive than EWC since it does not require computing statistics such as the Fisher information over the pre-training dataset.

\section{Conclusion}

In this work, we proposed a novel method for PET, Dynamic Subset Tuning (DST). DST is more flexible than extant PET methods, as it \textit{simultaneously} optimizes both the subset of parameters to update and their corresponding values, and dynamically re-selects that subset every training step. We showed that because the size of the updated subset can be arbitrarily set, smaller subsets can be trained than with other methods, yielding a smaller PET model size. We showed that DST matches or outperforms LoRA and prompt tuning on a suite of NLP tasks and across various model types and sizes.

Our appendix includes analysis that sheds light on the function and application of DST, investigating the impact of churn, siloing, subset generalization, and computational costs. The limitations of our work are listed in Appendix \ref{sec:limitations}.

\bibliographystyle{acl_natbib}
\bibliography{refs}

\begin{thebibliography}{51}
\providecommand{\natexlab}[1]{#1}

\bibitem[{Anil et~al.(2023)Anil, Dai, Firat, Johnson, Lepikhin, Passos,
  Shakeri, Taropa, Bailey, Chen et~al.}]{palm2}
Rohan Anil, Andrew~M Dai, Orhan Firat, Melvin Johnson, Dmitry Lepikhin,
  Alexandre Passos, Siamak Shakeri, Emanuel Taropa, Paige Bailey, Zhifeng Chen,
  et~al. 2023.
\newblock Palm 2 technical report.
\newblock \emph{arXiv preprint arXiv:2305.10403}.

\bibitem[{Ansell et~al.(2022)Ansell, Ponti, Korhonen, and
  Vuli{\'c}}]{ansell-etal-2022-composable}
Alan Ansell, Edoardo Ponti, Anna Korhonen, and Ivan Vuli{\'c}. 2022.
\newblock \href {https://doi.org/10.18653/v1/2022.acl-long.125} {Composable
  sparse fine-tuning for cross-lingual transfer}.
\newblock In \emph{Proceedings of the 60th Annual Meeting of the Association
  for Computational Linguistics (Volume 1: Long Papers)}, pages 1778--1796,
  Dublin, Ireland. Association for Computational Linguistics.

\bibitem[{Ben-Zaken et~al.(2022)Ben-Zaken, Goldberg, and
  Ravfogel}]{ben-zaken-etal-2022-bitfit}
Elad Ben-Zaken, Yoav Goldberg, and Shauli Ravfogel. 2022.
\newblock \href {https://doi.org/10.18653/v1/2022.acl-short.1} {{B}it{F}it:
  Simple parameter-efficient fine-tuning for transformer-based masked
  language-models}.
\newblock In \emph{Proceedings of the 60th Annual Meeting of the Association
  for Computational Linguistics (Volume 2: Short Papers)}, pages 1--9, Dublin,
  Ireland. Association for Computational Linguistics.

\bibitem[{Bradbury et~al.(2018)Bradbury, Frostig, Hawkins, Johnson, Leary,
  Maclaurin, Necula, Paszke, Vander{P}las, Wanderman-{M}ilne, and Zhang}]{jax}
James Bradbury, Roy Frostig, Peter Hawkins, Matthew~James Johnson, Chris Leary,
  Dougal Maclaurin, George Necula, Adam Paszke, Jake Vander{P}las, Skye
  Wanderman-{M}ilne, and Qiao Zhang. 2018.
\newblock \href {http://github.com/google/jax} {{JAX}: composable
  transformations of {P}ython+{N}um{P}y programs}.

\bibitem[{Chen et~al.(2022)Chen, Liu, Meng, and
  Liang}]{chen-etal-2022-revisiting}
Guanzheng Chen, Fangyu Liu, Zaiqiao Meng, and Shangsong Liang. 2022.
\newblock \href {https://doi.org/10.18653/v1/2022.emnlp-main.168} {Revisiting
  parameter-efficient tuning: Are we really there yet?}
\newblock In \emph{Proceedings of the 2022 Conference on Empirical Methods in
  Natural Language Processing}, pages 2612--2626, Abu Dhabi, United Arab
  Emirates. Association for Computational Linguistics.

\bibitem[{Chowdhery et~al.(2022)Chowdhery, Narang, Devlin, Bosma, Mishra,
  Roberts, Barham, Chung, Sutton, Gehrmann et~al.}]{palm}
Aakanksha Chowdhery, Sharan Narang, Jacob Devlin, Maarten Bosma, Gaurav Mishra,
  Adam Roberts, Paul Barham, Hyung~Won Chung, Charles Sutton, Sebastian
  Gehrmann, et~al. 2022.
\newblock Palm: Scaling language modeling with pathways.
\newblock \emph{arXiv preprint arXiv:2204.02311}.

\bibitem[{Cobbe et~al.(2021)Cobbe, Kosaraju, Bavarian, Chen, Jun, Kaiser,
  Plappert, Tworek, Hilton, Nakano et~al.}]{gsm8k}
Karl Cobbe, Vineet Kosaraju, Mohammad Bavarian, Mark Chen, Heewoo Jun, Lukasz
  Kaiser, Matthias Plappert, Jerry Tworek, Jacob Hilton, Reiichiro Nakano,
  et~al. 2021.
\newblock Training verifiers to solve math word problems.
\newblock \emph{arXiv preprint arXiv:2110.14168}.

\bibitem[{Dettmers et~al.(2021)Dettmers, Lewis, Shleifer, and
  Zettlemoyer}]{dettmers20218}
Tim Dettmers, Mike Lewis, Sam Shleifer, and Luke Zettlemoyer. 2021.
\newblock 8-bit optimizers via block-wise quantization.
\newblock \emph{arXiv preprint arXiv:2110.02861}.

\bibitem[{Ding et~al.(2022)Ding, Qin, Yang, Wei, Yang, Su, Hu, Chen, Chan, Chen
  et~al.}]{delta-tuning-overview}
Ning Ding, Yujia Qin, Guang Yang, Fuchao Wei, Zonghan Yang, Yusheng Su,
  Shengding Hu, Yulin Chen, Chi-Min Chan, Weize Chen, et~al. 2022.
\newblock Delta tuning: A comprehensive study of parameter efficient methods
  for pre-trained language models.
\newblock \emph{arXiv preprint arXiv:2203.06904}.

\bibitem[{Finn et~al.(2017)Finn, Abbeel, and Levine}]{pmlr-v70-finn17a}
Chelsea Finn, Pieter Abbeel, and Sergey Levine. 2017.
\newblock \href {https://proceedings.mlr.press/v70/finn17a.html}
  {Model-agnostic meta-learning for fast adaptation of deep networks}.
\newblock In \emph{Proceedings of the 34th International Conference on Machine
  Learning}, volume~70 of \emph{Proceedings of Machine Learning Research},
  pages 1126--1135. PMLR.

\bibitem[{Frankle and Carbin(2019)}]{lth}
Jonathan Frankle and Michael Carbin. 2019.
\newblock \href {https://openreview.net/forum?id=rJl-b3RcF7} {The lottery
  ticket hypothesis: Finding sparse, trainable neural networks}.
\newblock In \emph{International Conference on Learning Representations}.

\bibitem[{Gong et~al.(2019)Gong, He, Li, Qin, Wang, and Liu}]{stacking}
Linyuan Gong, Di~He, Zhuohan Li, Tao Qin, Liwei Wang, and Tieyan Liu. 2019.
\newblock \href {https://proceedings.mlr.press/v97/gong19a.html} {Efficient
  training of {BERT} by progressively stacking}.
\newblock In \emph{Proceedings of the 36th International Conference on Machine
  Learning}, volume~97 of \emph{Proceedings of Machine Learning Research},
  pages 2337--2346. PMLR.

\bibitem[{Guo et~al.(2021)Guo, Rush, and Kim}]{diff-pruning}
Demi Guo, Alexander Rush, and Yoon Kim. 2021.
\newblock \href {https://doi.org/10.18653/v1/2021.acl-long.378}
  {Parameter-efficient transfer learning with diff pruning}.
\newblock In \emph{Proceedings of the 59th Annual Meeting of the Association
  for Computational Linguistics and the 11th International Joint Conference on
  Natural Language Processing (Volume 1: Long Papers)}, pages 4884--4896,
  Online. Association for Computational Linguistics.

\bibitem[{Han et~al.(2022)Han, Wu, Hu, and Chen}]{han-etal-2022-lan}
Bing Han, Yangjian Wu, Gang Hu, and Qiulin Chen. 2022.
\newblock \href {https://aclanthology.org/2022.wmt-1.19} {Lan-bridge {MT}{'}s
  participation in the {WMT} 2022 general translation shared task}.
\newblock In \emph{Proceedings of the Seventh Conference on Machine Translation
  (WMT)}, pages 268--274, Abu Dhabi, United Arab Emirates (Hybrid). Association
  for Computational Linguistics.

\bibitem[{Han et~al.(2015)Han, Pool, Tran, and Dally}]{magnitude-pruning}
Song Han, Jeff Pool, John Tran, and William Dally. 2015.
\newblock \href
  {https://proceedings.neurips.cc/paper_files/paper/2015/file/ae0eb3eed39d2bcef4622b2499a05fe6-Paper.pdf}
  {Learning both weights and connections for efficient neural network}.
\newblock In \emph{Advances in Neural Information Processing Systems},
  volume~28. Curran Associates, Inc.

\bibitem[{He et~al.(2021)He, Liu, Ye, Tan, Ding, Cheng, Low, Bing, and
  Si}]{he-etal-2021-effectiveness}
Ruidan He, Linlin Liu, Hai Ye, Qingyu Tan, Bosheng Ding, Liying Cheng, Jiawei
  Low, Lidong Bing, and Luo Si. 2021.
\newblock \href {https://doi.org/10.18653/v1/2021.acl-long.172} {On the
  effectiveness of adapter-based tuning for pretrained language model
  adaptation}.
\newblock In \emph{Proceedings of the 59th Annual Meeting of the Association
  for Computational Linguistics and the 11th International Joint Conference on
  Natural Language Processing (Volume 1: Long Papers)}, pages 2208--2222,
  Online. Association for Computational Linguistics.

\bibitem[{Houlsby et~al.(2019)Houlsby, Giurgiu, Jastrzebski, Morrone,
  De~Laroussilhe, Gesmundo, Attariyan, and Gelly}]{adapter}
Neil Houlsby, Andrei Giurgiu, Stanislaw Jastrzebski, Bruna Morrone, Quentin
  De~Laroussilhe, Andrea Gesmundo, Mona Attariyan, and Sylvain Gelly. 2019.
\newblock \href {https://proceedings.mlr.press/v97/houlsby19a.html}
  {Parameter-efficient transfer learning for {NLP}}.
\newblock In \emph{Proceedings of the 36th International Conference on Machine
  Learning}, volume~97 of \emph{Proceedings of Machine Learning Research},
  pages 2790--2799. PMLR.

\bibitem[{Hu et~al.(2022)Hu, yelong shen, Wallis, Allen-Zhu, Li, Wang, Wang,
  and Chen}]{lora}
Edward~J Hu, yelong shen, Phillip Wallis, Zeyuan Allen-Zhu, Yuanzhi Li, Shean
  Wang, Lu~Wang, and Weizhu Chen. 2022.
\newblock \href {https://openreview.net/forum?id=nZeVKeeFYf9} {Lo{RA}: Low-rank
  adaptation of large language models}.
\newblock In \emph{International Conference on Learning Representations}.

\bibitem[{Joshi et~al.(2017)Joshi, Choi, Weld, and
  Zettlemoyer}]{joshi-etal-2017-triviaqa}
Mandar Joshi, Eunsol Choi, Daniel Weld, and Luke Zettlemoyer. 2017.
\newblock \href {https://doi.org/10.18653/v1/P17-1147} {{T}rivia{QA}: A large
  scale distantly supervised challenge dataset for reading comprehension}.
\newblock In \emph{Proceedings of the 55th Annual Meeting of the Association
  for Computational Linguistics (Volume 1: Long Papers)}, pages 1601--1611,
  Vancouver, Canada. Association for Computational Linguistics.

\bibitem[{Kirkpatrick et~al.(2017)Kirkpatrick, Pascanu, Rabinowitz, Veness,
  Desjardins, Rusu, Milan, Quan, Ramalho, Grabska-Barwinska et~al.}]{ewc}
James Kirkpatrick, Razvan Pascanu, Neil Rabinowitz, Joel Veness, Guillaume
  Desjardins, Andrei~A Rusu, Kieran Milan, John Quan, Tiago Ramalho, Agnieszka
  Grabska-Barwinska, et~al. 2017.
\newblock Overcoming catastrophic forgetting in neural networks.
\newblock \emph{Proceedings of the national academy of sciences},
  114(13):3521--3526.

\bibitem[{Kocmi et~al.(2022)Kocmi, Bawden, Bojar, Dvorkovich, Federmann,
  Fishel, Gowda, Graham, Grundkiewicz, Haddow, Knowles, Koehn, Monz, Morishita,
  Nagata, Nakazawa, Nov{\'a}k, Popel, and
  Popovi{\'c}}]{kocmi-etal-2022-findings}
Tom Kocmi, Rachel Bawden, Ond{\v{r}}ej Bojar, Anton Dvorkovich, Christian
  Federmann, Mark Fishel, Thamme Gowda, Yvette Graham, Roman Grundkiewicz,
  Barry Haddow, Rebecca Knowles, Philipp Koehn, Christof Monz, Makoto
  Morishita, Masaaki Nagata, Toshiaki Nakazawa, Michal Nov{\'a}k, Martin Popel,
  and Maja Popovi{\'c}. 2022.
\newblock \href {https://aclanthology.org/2022.wmt-1.1} {Findings of the 2022
  conference on machine translation ({WMT}22)}.
\newblock In \emph{Proceedings of the Seventh Conference on Machine Translation
  (WMT)}, pages 1--45, Abu Dhabi, United Arab Emirates (Hybrid). Association
  for Computational Linguistics.

\bibitem[{Koohpayegani et~al.(2024)Koohpayegani, Navaneet, Nooralinejad,
  Kolouri, and Pirsiavash}]{compressinglora}
Soroush~Abbasi Koohpayegani, KL~Navaneet, Parsa Nooralinejad, Soheil Kolouri,
  and Hamed Pirsiavash. 2024.
\newblock Nola: Compressing lora using linear combination of random basis.
\newblock In \emph{The Twelfth International Conference on Learning
  Representations (ICLR)}.

\bibitem[{Lee et~al.(2020)Lee, Cho, and Kang}]{mixout}
Cheolhyoung Lee, Kyunghyun Cho, and Wanmo Kang. 2020.
\newblock Mixout: Effective regularization to finetune large-scale pretrained
  language models.
\newblock In \emph{8th International Conference on Learning Representations,
  {ICLR} 2020, Addis Ababa, Ethiopia, April 26-30, 2020}.

\bibitem[{Lester et~al.(2021)Lester, Al-Rfou, and Constant}]{pt}
Brian Lester, Rami Al-Rfou, and Noah Constant. 2021.
\newblock \href {https://doi.org/10.18653/v1/2021.emnlp-main.243} {The power of
  scale for parameter-efficient prompt tuning}.
\newblock In \emph{Proceedings of the 2021 Conference on Empirical Methods in
  Natural Language Processing}, pages 3045--3059, Online and Punta Cana,
  Dominican Republic. Association for Computational Linguistics.

\bibitem[{Li et~al.(2023)Li, Chen, and Zhu}]{NEURIPS2023_3122aaa2}
Bingrui Li, Jianfei Chen, and Jun Zhu. 2023.
\newblock \href
  {https://proceedings.neurips.cc/paper_files/paper/2023/file/3122aaa22b2fe83f9cead1a696f65ceb-Paper-Conference.pdf}
  {Memory efficient optimizers with 4-bit states}.
\newblock In \emph{Advances in Neural Information Processing Systems},
  volume~36, pages 15136--15171. Curran Associates, Inc.

\bibitem[{Liu et~al.(2022)Liu, Tam, Muqeeth, Mohta, Huang, Bansal, and
  Raffel}]{NEURIPS2022_0cde695b}
Haokun Liu, Derek Tam, Mohammed Muqeeth, Jay Mohta, Tenghao Huang, Mohit
  Bansal, and Colin~A Raffel. 2022.
\newblock \href
  {https://proceedings.neurips.cc/paper_files/paper/2022/file/0cde695b83bd186c1fd456302888454c-Paper-Conference.pdf}
  {Few-shot parameter-efficient fine-tuning is better and cheaper than
  in-context learning}.
\newblock In \emph{Advances in Neural Information Processing Systems},
  volume~35, pages 1950--1965. Curran Associates, Inc.

\bibitem[{Mallya et~al.(2018)Mallya, Davis, and Lazebnik}]{piggyback}
Arun Mallya, Dillon Davis, and Svetlana Lazebnik. 2018.
\newblock Piggyback: Adapting a single network to multiple tasks by learning to
  mask weights.
\newblock In \emph{Proceedings of the European conference on computer vision
  (ECCV)}, pages 67--82.

\bibitem[{OpenAI(2023)}]{gpt4}
OpenAI. 2023.
\newblock {GPT-4} technical report.
\newblock \emph{arXiv preprint arXiv:2303.08774}.

\bibitem[{Oymak et~al.(2023)Oymak, Rawat, Soltanolkotabi, and
  Thrampoulidis}]{pmlr-v202-oymak23a}
Samet Oymak, Ankit~Singh Rawat, Mahdi Soltanolkotabi, and Christos
  Thrampoulidis. 2023.
\newblock \href {https://proceedings.mlr.press/v202/oymak23a.html} {On the role
  of attention in prompt-tuning}.
\newblock In \emph{Proceedings of the 40th International Conference on Machine
  Learning}, volume 202 of \emph{Proceedings of Machine Learning Research},
  pages 26724--26768. PMLR.

\bibitem[{Panigrahi et~al.(2023)Panigrahi, Saunshi, Zhao, and
  Arora}]{skilllocalization}
Abhishek Panigrahi, Nikunj Saunshi, Haoyu Zhao, and Sanjeev Arora. 2023.
\newblock \href {https://proceedings.mlr.press/v202/panigrahi23a.html}
  {Task-specific skill localization in fine-tuned language models}.
\newblock In \emph{Proceedings of the 40th International Conference on Machine
  Learning}, volume 202 of \emph{Proceedings of Machine Learning Research},
  pages 27011--27033. PMLR.

\bibitem[{Pfeiffer et~al.(2021)Pfeiffer, Kamath, R{\"u}ckl{\'e}, Cho, and
  Gurevych}]{pfeiffer-etal-2021-adapterfusion}
Jonas Pfeiffer, Aishwarya Kamath, Andreas R{\"u}ckl{\'e}, Kyunghyun Cho, and
  Iryna Gurevych. 2021.
\newblock \href {https://doi.org/10.18653/v1/2021.eacl-main.39}
  {{A}dapter{F}usion: Non-destructive task composition for transfer learning}.
\newblock In \emph{Proceedings of the 16th Conference of the European Chapter
  of the Association for Computational Linguistics: Main Volume}, pages
  487--503, Online. Association for Computational Linguistics.

\bibitem[{Post(2018)}]{post-2018-call}
Matt Post. 2018.
\newblock \href {https://www.aclweb.org/anthology/W18-6319} {A call for clarity
  in reporting {BLEU} scores}.
\newblock In \emph{Proceedings of the Third Conference on Machine Translation:
  Research Papers}, pages 186--191, Belgium, Brussels. Association for
  Computational Linguistics.

\bibitem[{Raffel et~al.(2020)Raffel, Shazeer, Roberts, Lee, Narang, Matena,
  Zhou, Li, and Liu}]{t5}
Colin Raffel, Noam Shazeer, Adam Roberts, Katherine Lee, Sharan Narang, Michael
  Matena, Yanqi Zhou, Wei Li, and Peter~J. Liu. 2020.
\newblock Exploring the limits of transfer learning with a unified text-to-text
  transformer.
\newblock \emph{J. Mach. Learn. Res.}, 21(1).

\bibitem[{Rebuffi et~al.(2017)Rebuffi, Bilen, and Vedaldi}]{NIPS2017_e7b24b11}
Sylvestre-Alvise Rebuffi, Hakan Bilen, and Andrea Vedaldi. 2017.
\newblock \href
  {https://proceedings.neurips.cc/paper_files/paper/2017/file/e7b24b112a44fdd9ee93bdf998c6ca0e-Paper.pdf}
  {Learning multiple visual domains with residual adapters}.
\newblock In \emph{Advances in Neural Information Processing Systems},
  volume~30. Curran Associates, Inc.

\bibitem[{Roberts et~al.(2022)Roberts, Chung, Levskaya, Mishra, Bradbury,
  Andor, Narang, Lester, Gaffney, Mohiuddin, Hawthorne, Lewkowycz, Salcianu,
  van Zee, Austin, Goodman, Soares, Hu, Tsvyashchenko, Chowdhery, Bastings,
  Bulian, Garcia, Ni, Chen, Kenealy, Clark, Lee, Garrette, Lee-Thorp, Raffel,
  Shazeer, Ritter, Bosma, Passos, Maitin-Shepard, Fiedel, Omernick, Saeta,
  Sepassi, Spiridonov, Newlan, and Gesmundo}]{roberts2022t5x}
Adam Roberts, Hyung~Won Chung, Anselm Levskaya, Gaurav Mishra, James Bradbury,
  Daniel Andor, Sharan Narang, Brian Lester, Colin Gaffney, Afroz Mohiuddin,
  Curtis Hawthorne, Aitor Lewkowycz, Alex Salcianu, Marc van Zee, Jacob Austin,
  Sebastian Goodman, Livio~Baldini Soares, Haitang Hu, Sasha Tsvyashchenko,
  Aakanksha Chowdhery, Jasmijn Bastings, Jannis Bulian, Xavier Garcia, Jianmo
  Ni, Andrew Chen, Kathleen Kenealy, Jonathan~H. Clark, Stephan Lee, Dan
  Garrette, James Lee-Thorp, Colin Raffel, Noam Shazeer, Marvin Ritter, Maarten
  Bosma, Alexandre Passos, Jeremy Maitin-Shepard, Noah Fiedel, Mark Omernick,
  Brennan Saeta, Ryan Sepassi, Alexander Spiridonov, Joshua Newlan, and Andrea
  Gesmundo. 2022.
\newblock \href {https://arxiv.org/abs/2203.17189} {Scaling up models and data
  with $\texttt{t5x}$ and $\texttt{seqio}$}.
\newblock \emph{arXiv preprint arXiv:2203.17189}.

\bibitem[{Roberts et~al.(2023)Roberts, Chung, Mishra, Levskaya, Bradbury,
  Andor, Narang, Lester, Gaffney, Mohiuddin, Hawthorne, Lewkowycz, Salcianu,
  van Zee, Austin, Goodman, Soares, Hu, Tsvyashchenko, Chowdhery, Bastings,
  Bulian, Garcia, Ni, Chen, Kenealy, Han, Casbon, Clark, Lee, Garrette,
  Lee-Thorp, Raffel, Shazeer, Ritter, Bosma, Passos, Maitin-Shepard, Fiedel,
  Omernick, Saeta, Sepassi, Spiridonov, Newlan, and Gesmundo}]{t5x}
Adam Roberts, Hyung~Won Chung, Gaurav Mishra, Anselm Levskaya, James Bradbury,
  Daniel Andor, Sharan Narang, Brian Lester, Colin Gaffney, Afroz Mohiuddin,
  Curtis Hawthorne, Aitor Lewkowycz, Alex Salcianu, Marc van Zee, Jacob Austin,
  Sebastian Goodman, Livio~Baldini Soares, Haitang Hu, Sasha Tsvyashchenko,
  Aakanksha Chowdhery, Jasmijn Bastings, Jannis Bulian, Xavier Garcia, Jianmo
  Ni, Andrew Chen, Kathleen Kenealy, Kehang Han, Michelle Casbon, Jonathan~H.
  Clark, Stephan Lee, Dan Garrette, James Lee-Thorp, Colin Raffel, Noam
  Shazeer, Marvin Ritter, Maarten Bosma, Alexandre Passos, Jeremy
  Maitin-Shepard, Noah Fiedel, Mark Omernick, Brennan Saeta, Ryan Sepassi,
  Alexander Spiridonov, Joshua Newlan, and Andrea Gesmundo. 2023.
\newblock \href {http://jmlr.org/papers/v24/23-0795.html} {Scaling up models
  and data with t5x and seqio}.
\newblock \emph{Journal of Machine Learning Research}, 24(377):1--8.

\bibitem[{Roberts et~al.(2020)Roberts, Raffel, and
  Shazeer}]{roberts-etal-2020-much}
Adam Roberts, Colin Raffel, and Noam Shazeer. 2020.
\newblock \href {https://doi.org/10.18653/v1/2020.emnlp-main.437} {How much
  knowledge can you pack into the parameters of a language model?}
\newblock In \emph{Proceedings of the 2020 Conference on Empirical Methods in
  Natural Language Processing (EMNLP)}, pages 5418--5426, Online. Association
  for Computational Linguistics.

\bibitem[{Sanh et~al.(2020)Sanh, Wolf, and Rush}]{movement-pruning}
Victor Sanh, Thomas Wolf, and Alexander Rush. 2020.
\newblock \href
  {https://proceedings.neurips.cc/paper_files/paper/2020/file/eae15aabaa768ae4a5993a8a4f4fa6e4-Paper.pdf}
  {Movement pruning: Adaptive sparsity by fine-tuning}.
\newblock In \emph{Advances in Neural Information Processing Systems},
  volume~33, pages 20378--20389. Curran Associates, Inc.

\bibitem[{Sellam et~al.(2020)Sellam, Das, and Parikh}]{sellam-etal-2020-bleurt}
Thibault Sellam, Dipanjan Das, and Ankur Parikh. 2020.
\newblock \href {https://doi.org/10.18653/v1/2020.acl-main.704} {{BLEURT}:
  Learning robust metrics for text generation}.
\newblock In \emph{Proceedings of the 58th Annual Meeting of the Association
  for Computational Linguistics}, pages 7881--7892, Online. Association for
  Computational Linguistics.

\bibitem[{Shazeer and Stern(2018)}]{adafactor}
Noam Shazeer and Mitchell Stern. 2018.
\newblock Adafactor: Adaptive learning rates with sublinear memory cost.
\newblock In \emph{International Conference on Machine Learning}, pages
  4596--4604. PMLR.

\bibitem[{Su et~al.(2023)Su, Chan, Cheng, Qin, Lin, Hu, Yang, Ding, Sun, Xie,
  Liu, and Sun}]{su-etal-2023-exploring}
Yusheng Su, Chi-Min Chan, Jiali Cheng, Yujia Qin, Yankai Lin, Shengding Hu,
  Zonghan Yang, Ning Ding, Xingzhi Sun, Guotong Xie, Zhiyuan Liu, and Maosong
  Sun. 2023.
\newblock \href {https://doi.org/10.18653/v1/2023.emnlp-main.931} {Exploring
  the impact of model scaling on parameter-efficient tuning}.
\newblock In \emph{Proceedings of the 2023 Conference on Empirical Methods in
  Natural Language Processing}, pages 15062--15078, Singapore. Association for
  Computational Linguistics.

\bibitem[{Sung et~al.(2021)Sung, Nair, and Raffel}]{fixed-mask}
Yi-Lin Sung, Varun Nair, and Colin~A Raffel. 2021.
\newblock Training neural networks with fixed sparse masks.
\newblock \emph{Advances in Neural Information Processing Systems},
  34:24193--24205.

\bibitem[{Vilar et~al.(2023)Vilar, Freitag, Cherry, Luo, Ratnakar, and
  Foster}]{vilar-etal-2023-prompting}
David Vilar, Markus Freitag, Colin Cherry, Jiaming Luo, Viresh Ratnakar, and
  George Foster. 2023.
\newblock \href {https://doi.org/10.18653/v1/2023.acl-long.859} {Prompting
  {P}a{LM} for translation: Assessing strategies and performance}.
\newblock In \emph{Proceedings of the 61st Annual Meeting of the Association
  for Computational Linguistics (Volume 1: Long Papers)}, pages 15406--15427,
  Toronto, Canada. Association for Computational Linguistics.

\bibitem[{Vu et~al.(2022)Vu, Lester, Constant, Al-Rfou{'}, and
  Cer}]{vu-etal-2022-spot}
Tu~Vu, Brian Lester, Noah Constant, Rami Al-Rfou{'}, and Daniel Cer. 2022.
\newblock \href {https://doi.org/10.18653/v1/2022.acl-long.346} {{SP}o{T}:
  Better frozen model adaptation through soft prompt transfer}.
\newblock In \emph{Proceedings of the 60th Annual Meeting of the Association
  for Computational Linguistics (Volume 1: Long Papers)}, pages 5039--5059,
  Dublin, Ireland. Association for Computational Linguistics.

\bibitem[{Wang et~al.(2019)Wang, Pruksachatkun, Nangia, Singh, Michael, Hill,
  Levy, and Bowman}]{wang2019superglue}
Alex Wang, Yada Pruksachatkun, Nikita Nangia, Amanpreet Singh, Julian Michael,
  Felix Hill, Omer Levy, and Samuel Bowman. 2019.
\newblock Superglue: A stickier benchmark for general-purpose language
  understanding systems.
\newblock \emph{Advances in neural information processing systems}, 32.

\bibitem[{Xu and Zhang(2024)}]{randommasking}
Jing Xu and Jingzhao Zhang. 2024.
\newblock Random masking finds winning tickets for parameter efficient
  fine-tuning.
\newblock \emph{arXiv preprint arXiv:2405.02596}.

\bibitem[{Xu et~al.(2021)Xu, Luo, Zhang, Tan, Chang, Huang, and
  Huang}]{child-tuning}
Runxin Xu, Fuli Luo, Zhiyuan Zhang, Chuanqi Tan, Baobao Chang, Songfang Huang,
  and Fei Huang. 2021.
\newblock \href {https://doi.org/10.18653/v1/2021.emnlp-main.749} {Raise a
  child in large language model: Towards effective and generalizable
  fine-tuning}.
\newblock In \emph{Proceedings of the 2021 Conference on Empirical Methods in
  Natural Language Processing}, pages 9514--9528, Online and Punta Cana,
  Dominican Republic. Association for Computational Linguistics.

\bibitem[{Zan et~al.(2022)Zan, Peng, Ding, Qiu, Liu, He, Lu, Zhang, Liu, Liu,
  Zhan, and Tao}]{zan-etal-2022-vega}
Changtong Zan, Keqin Peng, Liang Ding, Baopu Qiu, Boan Liu, Shwai He, Qingyu
  Lu, Zheng Zhang, Chuang Liu, Weifeng Liu, Yibing Zhan, and Dacheng Tao. 2022.
\newblock \href {https://aclanthology.org/2022.wmt-1.37} {Vega-{MT}: The {JD}
  explore academy machine translation system for {WMT}22}.
\newblock In \emph{Proceedings of the Seventh Conference on Machine Translation
  (WMT)}, pages 411--422, Abu Dhabi, United Arab Emirates (Hybrid). Association
  for Computational Linguistics.

\bibitem[{Zhang et~al.(2022{\natexlab{a}})Zhang, Li, Li, Zhang, Zhu, and
  Jin}]{dps}
Haojie Zhang, Ge~Li, Jia Li, Zhongjin Zhang, Yuqi Zhu, and Zhi Jin.
  2022{\natexlab{a}}.
\newblock Fine-tuning pre-trained language models effectively by optimizing
  subnetworks adaptively.
\newblock \emph{Advances in Neural Information Processing Systems},
  35:21442--21454.

\bibitem[{Zhang et~al.(2022{\natexlab{b}})Zhang, Lin, Liu, Li, Sun, and
  Zhou}]{zhang-etal-2022-moefication}
Zhengyan Zhang, Yankai Lin, Zhiyuan Liu, Peng Li, Maosong Sun, and Jie Zhou.
  2022{\natexlab{b}}.
\newblock \href {https://doi.org/10.18653/v1/2022.findings-acl.71}
  {{M}o{E}fication: Transformer feed-forward layers are mixtures of experts}.
\newblock In \emph{Findings of the Association for Computational Linguistics:
  ACL 2022}, pages 877--890, Dublin, Ireland. Association for Computational
  Linguistics.

\bibitem[{Zhao et~al.(2024)Zhao, Zhang, Chen, Wang, Anandkumar, and
  Tian}]{zhao2024galore}
Jiawei Zhao, Zhenyu Zhang, Beidi Chen, Zhangyang Wang, Anima Anandkumar, and
  Yuandong Tian. 2024.
\newblock \href {https://openreview.net/forum?id=hYHsrKDiX7} {Galore:
  Memory-efficient {LLM} training by gradient low-rank projection}.
\newblock In \emph{Forty-first International Conference on Machine Learning}.

\end{thebibliography}

%%%%%%%%%%%%%%%%%%%%%%%%%%%%%%%%%%%%%%%%%%%%%%%%%%%%%%%%%%%%

\clearpage

\appendix

\section{Distance functions and siloing}
\label{sec:dist-and-silo}

\begin{table}
\centering
\small
% \begin{tabular}{p{2in}l}
\begin{tabular}{ccc}
\hline 
\textbf{Absolute} & \textbf{Relative} & \textbf{Inverse-relative} \\
\hline
$d_\text{abs}(\theta_1, \theta_2, \theta^{(0)})=|\theta_1-\theta_2|$  & $d_\text{rel}(\theta_1, \theta_2, \theta^{(0)})=\big|\frac{\theta_1-\theta_2}{\theta^{(0)}}\big|$  & $d_\text{inv-rel}(\theta_1, \theta_2, \theta^{(0)})=|\theta^{(0)}(\theta_1-\theta_2)|$ \\
\hline
\end{tabular}
\caption{\label{tab:d} Parameter-wise distance functions used in this work to measure the difference between two models ($\theta_1$ and $\theta_2$) relative to the seed model ($\theta^{(0)}$).}
\end{table}

We used the \textbf{inverse-relative distance function} and \textbf{per-module-and-layer siloing} in the main paper. We experimented with the following alternative siloing strategies:

\begin{itemize}
    \item \textbf{No siloing}: The subset can be distributed freely across the full model.
    \item \textbf{Per-module siloing}: Each module in the network (attention, feed-forward, embeddings, etc.) receives the same fraction of free parameters, but the set can spread out across layers.
\end{itemize}

Furthermore, we explored different variants of the inverse-relative distance function (Table \ref{tab:d}).

\begin{table}
\centering
\small
\begin{tabular}{lc|c}
\hline
&  \textbf{$\epsilon=10^{-6}$} & \textbf{$\epsilon=10^{-4}$} \\
\hline
\multicolumn{3}{l}{\textbf{No siloing}} \\
\hline
Random & 71.45\% & 73.92\% \\
Absolute -- $d_\text{abs}(\cdot)$ & 71.20\% & 73.25\% \\
Relative -- $d_\text{rel}(\cdot)$ & 72.45\% & 74.04\% \\
Inverse-relative\ -- $d_\text{inv-rel}(\cdot)$ & 73.47\% & 74.06\% \\
\hline
\multicolumn{3}{l}{\textbf{Per-module siloing}} \\
\hline
Relative -- $d_\text{rel}(\cdot)$ & 72.11\% & 74.08\% \\
Inverse-relative\ -- $d_\text{inv-rel}(\cdot)$ & 73.30\% & 73.78\% \\
\hline
\multicolumn{3}{l}{\textbf{Per-module-and-layer siloing}} \\
\hline
Relative -- $d_\text{rel}(\cdot)$ & 72.68\% & 74.18\% \\
Inverse-relative\ -- $d_\text{inv-rel}(\cdot)$ & 73.83\% & 74.00\% \\
\hline
\end{tabular}
\caption{\label{tab:diff} MT development set performance of the Gecko model in terms of BLEURT for different siloing and scoring strategies.}
\end{table}

Table \ref{tab:diff} compares DST performance using the various distance functions and siloing strategies. We confirm the findings of \citet{su-etal-2023-exploring} and \citet{randommasking} that even random parameter selection\footnote{The random baseline randomly selects the free parameters once at the start of fine-tuning.} works well for sufficiently large subsets ($\epsilon=10^{-4}$). However, with $\epsilon=10^{-6}$ (i.e.\ only 0.0001\% of model parameters are free), the relative $d_\text{rel}(\cdot)$ distance and inverse-relative $d_\text{inv-rel}(\cdot)$ distance functions substantially outperform the random baseline. Per-module-and-layer siloing yields further gains. The BLEURT score differences between the methods are smaller for larger subsets ($\epsilon=10^{-4}$).

\paragraph{Training stability}

\begin{figure}[t!]
\centering
\small
\includegraphics[scale=0.95]{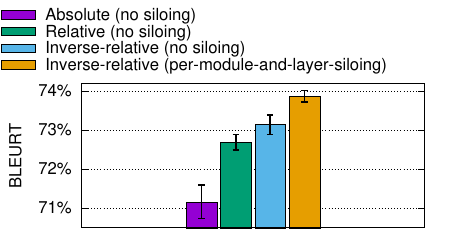}
\includegraphics[trim={0 0.5cm 0 0},clip,scale=0.95]{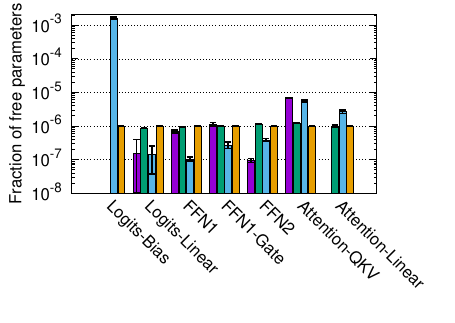}
\caption{BLEURT scores and the fraction of free parameters per network module on the MT development set (Gecko model, $\epsilon=10^{-6}$). Mean and standard deviation (error bars) across four training runs are shown.}
\label{fig:training-variation}
\end{figure}

\citet{chen-etal-2022-revisiting} demonstrated that training with existing PET methods like LoRA and prompt tuning is often unstable. The top panel in Fig.\ \ref{fig:training-variation} shows that the variance between training runs in DST depends on both distance function and siloing. Siloing and distance normalization (Relative and Inverse-relative) increase the average BLEURT and also reduce the variance compared to the absolute distance function (purple bar). The distribution of the free parameters across the network modules is most stable with the Relative distance function (green bar in the bottom panel in Fig.\ \ref{fig:training-variation}).

\section{Siloing alternatives}
\label{sec:silo-alt}

Siloing leads to the subset of free parameters being more evenly distributed across the full model, and thereby reduces the variability between training runs. We investigated two alternatives to siloing that aim to make $d(\cdot)$ more comparable across different modules:
\begin{itemize}
    \item Size normalization: Divide $d(\cdot)$ by the number of parameters in the module/layer.
    \item Mean normalization: Divide $d(\cdot)$ by the mean of the absolute seed values in the module/layer.
\end{itemize}

Table \ref{tab:silo-vs-norm} shows that siloing works more reliably than both strategies, especially with small $\epsilon$.

\begin{table*}[t!]
\centering
\small
\begin{tabular}{lcccc}
\hline 
\textbf{Method} & \multicolumn{2}{c}{\textbf{Per module}} & \multicolumn{2}{c}{\textbf{Per module and layer}} \\
 &  \textbf{$\epsilon=10^{-6}$} & \textbf{$\epsilon=10^{-4}$} &  \textbf{$\epsilon=10^{-6}$} & \textbf{$\epsilon=10^{-4}$} \\
\hline
Size normalization & 72.66\% & 73.84\% & 73.19\% & 73.79\% \\
Mean normalization & 71.23\% & 73.25\% & 70.80\% & 73.76\% \\
Siloing & 73.30\% & 73.78\% & 73.83\% & 74.00\% \\
\hline
\end{tabular}
\caption{\label{tab:silo-vs-norm} MT performance (BLEURT) of the Gecko model for alternatives to siloing.}
\end{table*}

\section{Efficient quantile computation}
\label{sec:quantile-computation}

The DST update step (Alg.\ \ref{alg:sdt}) involves computing the $q_S$ threshold as the $(1-\epsilon)$-th quantile over the difference vector $\Delta\big|_S$. The standard implementation in JAX (\texttt{jax.numpy.quantile()}) is based on sorting the complete $\Delta$ vector which can be slow for large models. In fact, \texttt{jax.numpy.quantile()} does not support tensor sizes beyond $2^{32}$, and is therefore not suitable for our larger models.

Fortunately, we noted that $q_S$ does not change abruptly between training steps but rather fluctuates smoothly over time. Therefore, we use an iterative approach that refines the threshold from the previous training step ($q'$) for a fixed number $m$ of iterations. The algorithm maintains a lower and an upper bound that are initialized with $[\frac{q'}{r},rq']$, i.e.\ we assume that $q_S$ does not increase/decrease by more than a factor of $r$ between training steps. Let $c(\cdot)$ be a function that counts the fraction of components in $\Delta$ greater than $q$:
\begin{equation}
    c(\Delta,q)=\frac{1}{|\Delta|}\sum_{i=1}^{|\Delta|}I(\Delta_i>q).
\end{equation}
The algorithm refines the lower and upper bounds for $m$ steps by iteratively setting one of the bounds to the midpoint $q_{mi}$ of the interval, depending on whether $c(\Delta, q_{mi})$ is less or greater than $\epsilon$. Alg.\ \ref{alg:iter-q} lists the full procedure. We found that the iterative method is able to track $q_S$ very closely with $m=3$ and $r=2$. After an initial ramp up period of around 20 training steps, the relative error between the approximated $\Tilde{\epsilon}$ (realized by $\Tilde{q}$) and the exact value of $\epsilon$ usually stays below 1\%.

\begin{algorithm}[t!]
\small
\caption{\small Iterative refinement algorithm for approximating the threshold $q_S$ for a single silo $S\in\mathcal{S}$.}
\label{alg:iter-q}
\begin{algorithmic}[1]
\REQUIRE{$q'$: The $q_S$ threshold from the previous training step.}
\REQUIRE{$\Delta$: Difference vector for $S$ of the current training step (see line 3 in Algorithm \ref{alg:sdt}).}
\REQUIRE{$\epsilon$: Fraction of free parameters.}
\REQUIRE{$m$: Number of refinement iterations.}
\REQUIRE{$r$: Maximum fluctuation factor.}
\STATE{$q_{lo} \gets \frac{q'}{r}$}
\STATE{$q_{mi} \gets q'$}
\STATE{$q_{hi} \gets rq'$}
\FOR{$k\gets 1$ \TO $m$}
\IF{$c(\Delta, q_{mi})>\epsilon$}
\STATE{$q_{lo} \gets q_{mi}$}
\ELSE
\STATE{$q_{hi} \gets q_{mi}$}
\ENDIF
\STATE{$q_{mi} \gets \frac{q_{hi}-q_{lo}}{2}$}
\ENDFOR
\RETURN{$\begin{cases}
			q_{lo}, & \text{if }|c(\Delta, q_{lo})-\epsilon|<|c(\Delta, q_{hi})-\epsilon| \\
            q_{hi}, & \text{otherwise}
		 \end{cases}$}
\end{algorithmic}
\end{algorithm}

\paragraph{Training speed}

\begin{figure}[t!]
\centering
\small
\includegraphics[scale=0.95]{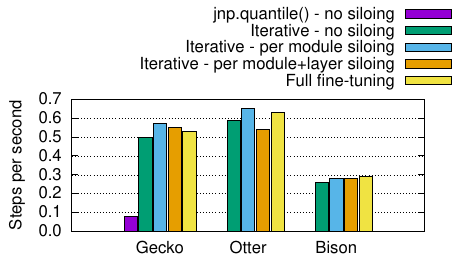}
\caption{Training steps per second on TPU v4 chips with a batch size of 32. We use a 2x2x4 topology for the Gecko model and 4x4x4 for Otter and Bison.}
\label{fig:steps-per-sec}
\end{figure}

Fig.\ \ref{fig:steps-per-sec} shows that a trivial implementation of DST using \texttt{jnp.quantile()} is much slower than full fine-tuning. However, using our novel approximate iterative algorithm to compute thresholds (Algorithm \ref{alg:iter-q}), we are able to match the full fine-tuning training speed. Algorithm \ref{alg:iter-q} is crucial for making dynamic subset tuning with churn possible in practice even for large models (Bison).

\begin{figure}[t!]
\centering
\small
\begin{tabular}{@{\hspace{0em}}c@{\hspace{0em}}c@{\hspace{0em}}}
\includegraphics[scale=0.21]{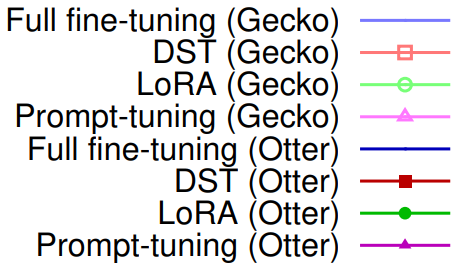} & \includegraphics[scale=0.9]{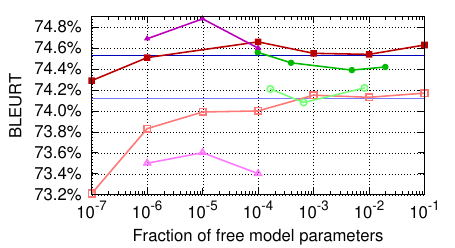} \\
& (a) Machine translation (German-English. WMT) \\
 \\
\includegraphics[scale=0.9]{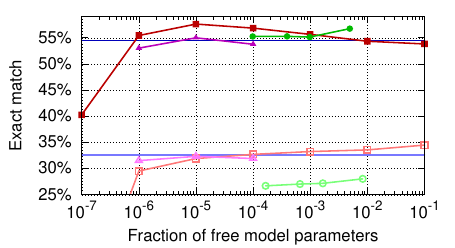} & \includegraphics[scale=0.9]{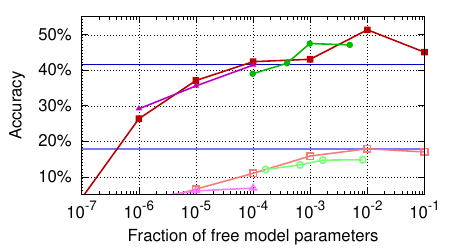}  \\
(b) TriviaQA & (c) GSM8K \\
\\
\includegraphics[scale=0.9]{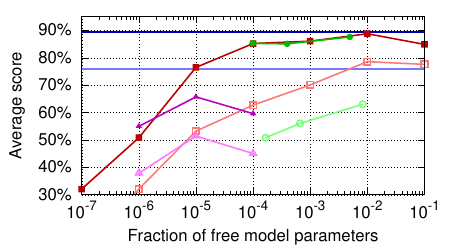} & \includegraphics[scale=0.9]{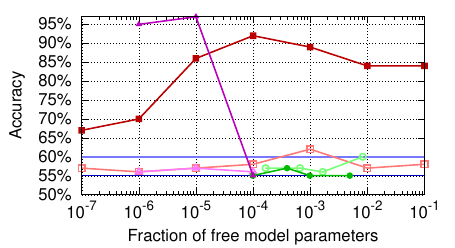} \\
(d) SuperGLUE & (e) CoPA \\
\end{tabular}

\caption{Validation set performance of the Gecko and Otter models as a function of the fraction of free parameters relative to full fine-tuning.}
\label{fig:main-eval-dev}
\end{figure}

\section{Absolute performance metrics}
\label{sec:main-eval-abs}

Fig.\ \ref{fig:main-eval-dev-rel} plots the performance curves of DST, LoRA, and prompt tuning relative to the respective fully fine-tuned baseline. Fig. \ref{fig:main-eval-dev} shows the same curves in absolute terms, marking the full fine-tuning baselines with blue (light: Gecko, dark: Otter) horizontal lines. The dark data series (Otter) outperform their lightly colored counter-parts (Gecko) because the Gecko model is 8x smaller than the Otter model.

The average scores on SuperGLUE for the fully fine-tuned T5 baselines are listed in Table~\ref{tab:t5-superglue}.

Table \ref{tab:wmt-comparison} shows that the BLEURT scores of our systems are comparable to the winners of the WMT22 evaluation campaign even though we used a very simple training pipeline (single fine-tuning stage on only 31.5K examples).

Table \ref{tab:wmt-test-bleu-bleurt} shows the absolute BLEU and BLEURT scores on the WMT22 German-English test set for all PET techniques used in this work.

\begin{table}[t!]
\centering
\small
\begin{minipage}{0.4\linewidth}
\centering
\small
\begin{tabular}{lc}
\hline
\textbf{T5 model} & \textbf{Avg.\ score} \\
\hline
T5-Small & 86.3\% \\
T5-Base & 88.5\% \\
T5-Large & 89.9\% \\
T5-XL & 91.6\% \\
T5-XXL & 92.3\% \\
\hline
\end{tabular}
\caption{\label{tab:t5-superglue} Average scores on SuperGLUE of the fully fine-tuned T5 baselines.}
\end{minipage}\hspace{0.6cm}
\begin{minipage}{0.55\linewidth}
\centering
\small
% \begin{tabular}{l@{\hspace{0.6em}}c@{\hspace{0.9em}}c}
\begin{tabular}{lcc}
\hline
& \textbf{BLEU} & \textbf{BLEURT} \\
\hline
JDExploreA.\ \citep{zan-etal-2022-vega}  & 49.3\% &	74.37\% \\
Online-B  & 49.7\%	& 74.00\% \\
Lan-Bridge \citep{han-etal-2022-lan}  & 50.1\% &	73.84\% \\
Online-A & 50.2\% & 73.08\% \\
\hline
DST (Bison model, $\epsilon=10^{-6}$) & 49.0\% &	74.13\%  \\
\hline
\end{tabular}
\caption{\label{tab:wmt-comparison} Comparison with official German-English WMT22 evaluation systems \citep{kocmi-etal-2022-findings}.}
\end{minipage}
\end{table}

\begin{table*}[t!]
\centering
\small
\begin{tabular}{l|cccc|ccc}
\hline
\textbf{Base} & \textbf{0-shot} & \textbf{Full fine-tuning} & \textbf{Prompt-tuning} & \textbf{LoRA} & \multicolumn{3}{c}{\textbf{This work (DST)}} \\
& & & \footnotesize  (length=10) & \footnotesize  (rank=1) & $\epsilon=10^{-7}$ & $\epsilon=10^{-6}$ & $\epsilon=10^{-4}$ \\
\hline
\multicolumn{8}{c}{\cellcolor{gray!10} \textbf{BLEU}} \\
\hline
Gecko & 47.0\% & 47.0\% & 43.8\% & 47.7\% & 46.8\% & 47.7\% & 47.2\% \\
Otter & 46.5\% & 47.3\% & 49.9\% & 46.9\% & 49.3\% & 49.5\% & 49.1\% \\
Bison & 47.9\% & 48.8\% & 48.4\% & 48.8\% & 49.5\% & 49.0\% & 48.4\% \\
\hline
\textbf{Avg.} & \textbf{47.1\%} & \textbf{47.7\%} & \textbf{47.4\%} & \textbf{47.8\%} & \textbf{48.5\%} & \textbf{48.7\%} & \textbf{48.2\%} \\
\hline
\multicolumn{8}{c}{\cellcolor{gray!10} \textbf{BLEURT}} \\
\hline
Gecko & 71.39\% & 73.41\% & 73.01\% & 73.35\% & 72.23\% & 73.08\% & 73.18\% \\
Otter & 69.96\% & 73.56\% & 73.92\% & 73.53\% & 73.48\% & 73.73\% & 74.15\% \\
Bison & 72.70\% & 74.11\% & 74.18\% & 74.17\% & 73.90\% & 74.13\% & 74.21\% \\
\hline
\textbf{Avg.} & \textbf{71.35\%} & \textbf{73.69\%} & \textbf{73.70\%} & \textbf{73.68\%} & \textbf{73.20\%} & \textbf{73.65\%} & \textbf{73.84\%} \\
\hline
\end{tabular}
\caption{\label{tab:wmt-test-bleu-bleurt} BLEU and BLEURT scores of PET techniques on the WMT22 German-English test set.}
\end{table*}

\section{Subset overlap analyses}
\label{sec:overlap-ana}

\paragraph{Churn}

\begin{table}[t!]
\centering
\small
\begin{tabular}{llcc}
\hline \textbf{$\epsilon$} & \textbf{Siloing} & \textbf{No churn} & \textbf{With churn} \\
\hline
& None & 71.38\% & 73.47\% \\
$10^{-6}$ & Module & 71.45\% & 73.30\% \\
 & Module+layer & 72.55\% & 73.83\% \\
\hline
 & None & 74.05\% & 74.06\% \\
$10^{-4}$ & Module & 73.50\% & 73.78\% \\
 & Module+layer & 73.69\% & 74.00\% \\
\hline
\end{tabular}
\caption{\label{tab:fixed-set} MT performance (BLEURT) of the Gecko model with and without subset churn.}
\end{table}

A core difference between DST and other delta tuning methods is that it re-selects the parameter subset in each training step, i.e.\ it allows subset churn. In an ablation study we disabled churn by selecting the subset only once in the first training step. However, as shown in Table\ \ref{tab:fixed-set}, churn helps the model to converge to a better subset, particularly when the subset is small ($\epsilon =10^{-6}$).

\begin{figure}[t!]
\centering
\small
\begin{tabular}{@{\hspace{0em}}|@{\hspace{0em}}c@{\hspace{0em}}|@{\hspace{0em}}c@{\hspace{0em}}|@{\hspace{0em}}c@{\hspace{0em}}|@{\hspace{0em}}}
\hline
& $\epsilon=10^{-4}$ & $\epsilon=10^{-6}$ \\
\hline
\rotatebox{90}{\hspace{5em}\textbf{No siloing}} & \includegraphics[scale=0.94]{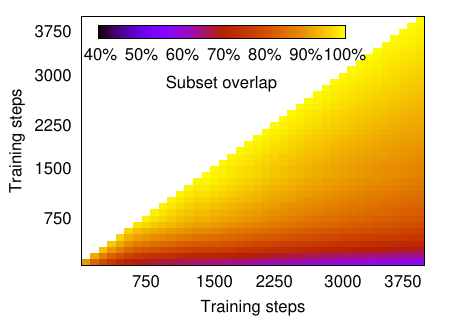} & \includegraphics[scale=0.94]{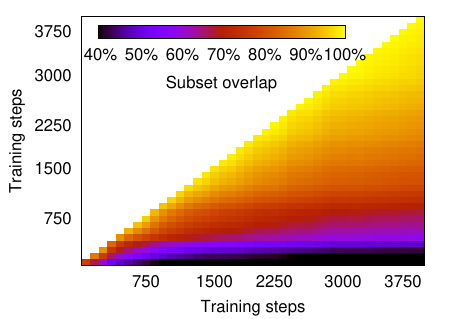} \\
\hline
\rotatebox{90}{\hspace{1.5em}\textbf{Per-module-and-layer siloing}} & \includegraphics[scale=0.94]{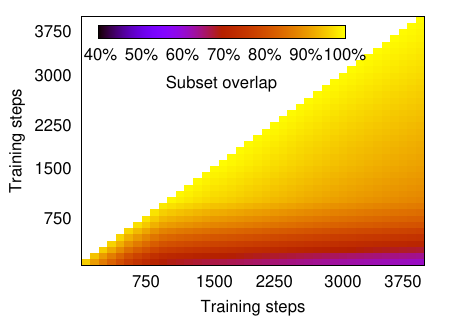} & \includegraphics[scale=0.94]{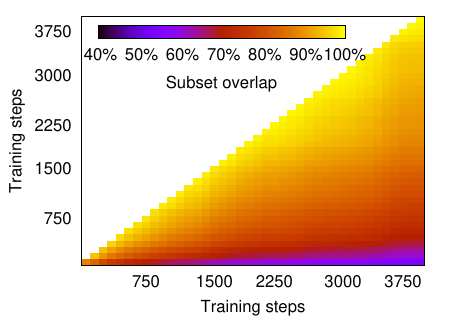} \\
\hline
\end{tabular}
\caption{Subset overlaps between different training steps of the same training run for the Gecko model on MT.}
\label{fig:churn-matrix}
\end{figure}

To illustrate how the subset changes in the course of training we compute the {\em subset overlap} between two DST checkpoints as follows:
\begin{equation}
    SubsetOverlap = \frac{|SubsetA \cap SubsetB|}{|SubsetA|}.
\end{equation}
Fig.\ \ref{fig:churn-matrix} shows that the subset tends to converge slower without siloing and with a small $\epsilon$. The overlap between the earliest and the latest checkpoint (bottom right corner in each diagram) is below 60\% which indicates the importance of churn.

\paragraph{Subset generalization}

\begin{figure}[t!]
\centering
\small
\begin{tabular}{@{\hspace{0em}}c@{\hspace{0em}}c@{\hspace{0em}}}
\includegraphics[trim={1.6cm 0.5cm 0 0},clip,scale=0.95]{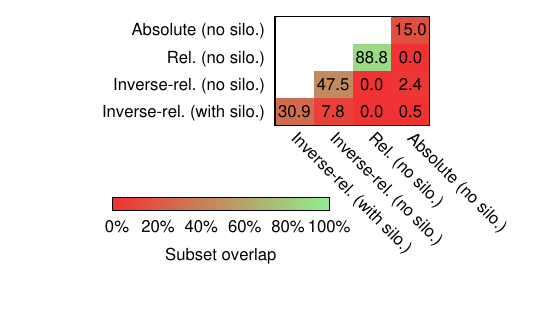} & \includegraphics[scale=0.95]{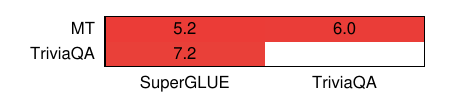} \\
(a) Overlap (\%) between different distance functions  & (b) Overlap (\%) between different tasks\\
\end{tabular}
\caption{Subset overlap averaged over four training runs (Gecko model, $\epsilon=10^{-6}$, 3K training steps).}
\label{fig:overlap}
\end{figure}

A practical question is how well the learnt subset masks generalize to new tasks. The off-diagonal values in Fig.\ \ref{fig:overlap}a indicate that there is very little subset overlap between different distance functions. The overlap stays below 50\% even between training runs of the same configuration (diagonal in Fig.\ \ref{fig:overlap}a) except for the Relative distance function. Subset overlaps between different tasks are relatively low, even when using the same distance function (Fig.\ \ref{fig:overlap}b). This suggests that LLMs are highly redundant: they contain many different parameter subsets, each sufficient to adapt the model to a certain task. Combining DST with approaches like MAML \citep{pmlr-v70-finn17a} could potentially improve the generalization and make it possible to re-use subset masks across tasks.
%, which would greatly reduce the memory footprint.

\section{Training hyper-parameters}
\label{sec:training-hp}

\begin{figure}[t!]
\centering
\small
\includegraphics[scale=0.95]{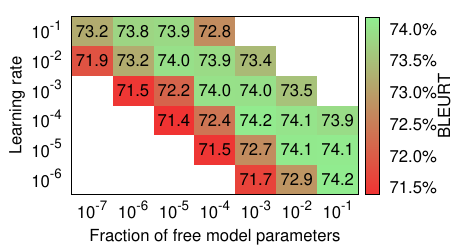}
\caption{MT performance (BLEURT) as a function of the learning rate and the number of free parameters ($\epsilon$).}
\label{fig:eps-lr}
\end{figure}

We train our models in the JAX \citep{jax} framework PAXML\footnote{\url{https://github.com/google/paxml}} on TPU v4 chips with AdaFactor \citep{adafactor}. We use a batch size of 32 for the PaLM 2 models and a batch size of 128 for T5. Dropout rates and learning rates are tuned for each experiment on the development sets. We used dropout for LoRA and full fine-tuning, but not for DST. We do not use dropout on GSM8K. For full fine-tuning, learning rates range between 0.1 and 1.0 in T5 and between 0.0001 and 0.000001 in PaLM 2.

Fig.\ \ref{fig:eps-lr} visualizes the inverse relationship between the optimal learning rate and $\epsilon$ in DST: the smaller the subset (i.e.\ the lower $\epsilon$), the larger the optimal learning rate. Intuitively, a low $\epsilon$ resets a larger fraction of the parameters to the seed model, and thus reduces the total magnitude of the update step. A higher learning rate compensates for this effect.
To reduce hallucinations in the PaLM 2 models we append the string ``\texttt{\textbackslash n[eod]}'' to all training examples and truncate predictions after these tokens.

\section{Datasets and evaluation protocols}
\label{sec:data}

 We test our method on the following benchmarks: Machine Translation (WMT22 German-English), Trivia-QA (closed-book, open-domain question answering), GSM8K (math problems), SuperGLUE (eight language understanding tasks). Appendix \ref{sec:data} lists more details about the datasets and our evaluation protocols.

\begin{itemize}
    \item {\bf Machine translation (MT)}. LLMs such as PaLM have been successfully applied to MT \citep{vilar-etal-2023-prompting}. Our main evaluation task is German-English translation using the official data from the WMT22 \citep{kocmi-etal-2022-findings} evaluation campaign. We use \texttt{newstest2022} as the test set to avoid overlap with PaLM training data, a 500 sentence sample from \texttt{newstest2021} as the development set, and all previous WMT News test sets as the fine-tuning set. We report BLEURT\footnote{\texttt{BLEURT-20} checkpoint} \citep{sellam-etal-2020-bleurt} scores against reference A and BLEU scores computed with SacreBLEU \citep{post-2018-call} against all available references.
    \item {\bf Trivia-QA (closed-book, open-domain)} \citep{roberts-etal-2020-much} is a version of the question answering task Trivia-QA \citep{joshi-etal-2017-triviaqa} without direct access to the context of the question. We follow the seqio \citep{roberts2022t5x} setup\footnote{\url{https://github.com/google/seqio\#triviaqa-closed-book-open-domain-version}} and report the number of exact matches.
    \item {\bf GSM8K} \citep{gsm8k} is a small dataset of high-quality grade school math word problems often used to assess LLMs.
    \item {\bf SuperGLUE} \citep{wang2019superglue} is a benchmark consisting of eight different language understanding tasks. Training set sizes range from 250 to 101K examples which we mix proportionally, following the \texttt{super\_glue\_v102\_proportional}\footnote{\url{https://github.com/google-research/text-to-text-transfer-transformer/blob/main/t5/data/mixtures.py}} recipe in T5 \citep{t5}. Like prior work we report average scores across all tasks, averaging scores of tasks with multiple scores first.
\end{itemize}

\begin{table}[t!]
\centering
\small
\begin{tabular}{lll}
\hline
\multicolumn{2}{l}{\textbf{Dataset}} & \textbf{Size} \\
\hline
\multicolumn{2}{l}{WMT German-English} & 31.5K \\
\hline
\multicolumn{2}{l}{Trivia-QA} & 87.6K \\
\hline
\multicolumn{2}{l}{GSM8K} & 7.5K \\
\hline
SuperGLUE & BoolQ & 9.4K \\
 & CB & 250 \\
 & COPA & 400 \\
 & MultiRC & 5.1K \\
 & ReCoRD & 101K \\
 & RTE & 2.5K \\
 & WiC & 6K \\
 & WSC & 554 \\
\hline
\end{tabular}
\caption{\label{tab:training-dataset-sizes} Number of examples in the training corpora.}
\end{table}

Table \ref{tab:training-dataset-sizes} lists the training data set sizes. We decode with beam search (beam size of 4) for MT, and with greedy search for the other benchmarks.

\section{Limitations}
\label{sec:limitations}

DST in its current form improves storage and regularization efficiency, but it does not improve the training efficiency or memory footprint. How to reduce the memory requirements in LLM adaptation is still an active area of research \citep{dettmers20218,NEURIPS2023_3122aaa2,zhao2024galore}.

% TODO: Mention that we want to explore other aspects of training with DST, e.g. catastrophic forgetting (adapters help: \citep{he-etal-2021-effectiveness})

DST is a new approach for PET. Similar to existing PET methods such as prompt tuning and LoRA, our approach may be potentially used in a setting which can propagate or amplify existing biases from the dataset.

We tested DST with PaLM 2 and T5 models of different sizes, but we did not use other LLMs due to license restrictions and computational costs. Furthermore, we leave it to future work to explore the combination of multiple PET techniques.

\end{document}